\documentclass[letterpaper, 10 pt, conference]{ieeeconf} 

\IEEEoverridecommandlockouts                        

                            \usepackage{tikz}
\usetikzlibrary{shapes.geometric, positioning, calc}
\usepackage{graphics}
\usepackage{amsmath} 
\usepackage{amssymb}
\usepackage{cite}
\usepackage{tikz}
\usepackage{amsmath}
\usepackage{tabularx}
\usepackage{array}
\usepackage{booktabs}
\usepackage{bm}
\usepackage{algorithm}
\usepackage{algorithmic}
\usepackage{bbm}
\usepackage{caption}
\PassOptionsToPackage{hyphens}{url}\usepackage{hyperref}

\usepackage{tablefootnote}

\DeclareMathOperator*{\argmin}{argmin}

\definecolor{green}{RGB}{11,155,13}

\title{\LARGE \bf
A Study on Learning Social Robot Navigation \\with Multimodal Perception
}

\author{Bhabaranjan Panigrahi, Amir Hossain Raj, Mohammad Nazeri, and Xuesu Xiao
\thanks{All authors are with the Department of Computer Science, George Mason University {\tt\scriptsize \{bpanigr, araj20, mnazerir, xiao\}@gmu.edu}}
}

\begin{document}
\maketitle
\thispagestyle{empty}
\pagestyle{empty}

\begin{abstract}
Autonomous mobile robots need to perceive the environments with their onboard sensors (e.g., LiDARs and RGB cameras) and then make appropriate navigation decisions. In order to navigate human-inhabited public spaces, such a navigation task becomes more than only obstacle avoidance, but also requires considering surrounding humans and their intentions to somewhat change the navigation behavior in response to the underlying social norms, i.e., being socially compliant. 
Machine learning methods are shown to be effective in capturing those complex and subtle social interactions in a data-driven manner, without explicitly hand-crafting simplified models or cost functions. 
Considering multiple available sensor modalities and the efficiency of learning methods, this paper presents a comprehensive study on learning social robot navigation with multimodal perception using a large-scale real-world dataset. The study investigates social robot navigation decision making on both the global and local planning levels and contrasts unimodal and multimodal learning against a set of classical navigation approaches in different social scenarios, while also analyzing the training and generalizability performance from the learning perspective. We also conduct a human study on how learning with multimodal perception affects the perceived social compliance. The results show that multimodal learning has a clear advantage over unimodal learning in both dataset and human studies. We open-source our code for the community's future use to study multimodal perception for learning social robot navigation.\footnote{GitHub: \url{https://github.com/RobotiXX/multimodal-fusion-network/}}

\end{abstract}

\section{INTRODUCTION}
\label{sec::introduction}

Thanks to decades of robotics research~\cite{fox1997dynamic, quinlan1993elastic}, autonomous mobile robots can navigate from one point to another in a collision-free manner in many real-world environments, e.g., factories and warehouses. Using onboard sensors, e.g., LiDARs and RGB cameras, those robots can perceive the environments, divide their workspaces into obstacles and free spaces, and then make navigation decisions to avoid obstacles and move towards their goal~\cite{perille2020benchmarking, nair2022dynabarn, xiao2022autonomous, xiao2023autonomous}. 

However, when deploying mobile robots in human-inhabited public spaces, the navigation task becomes more complex~\cite{mavrogiannis2023core, mirsky2021conflict, francis2023principles}: While avoiding any obstacle on the way to the goal is still required, they also need to consider other humans sharing the same environments and adjust their decision-making process to produce new navigation behaviors that respond to the underlying, usually unwritten, social norms. 

One avenue to achieve such social compliance is machine learning~\cite{xiao2022motion}. Learning approaches allow those complex and subtle human-robot interactions during social navigation to be captured in a data-driven manner and alleviate roboticists from manually designing simplified models~\cite{helbing1995social, van2011reciprocal}, crafting cost functions~\cite{xiao2022learning, kretzschmar2016socially}, and fine-tuning system parameters~\cite{xiao2022appl, xiao2020appld, wang2021apple, wang2021appli, xu2021applr}. The development of machine learning infrastructure, e.g., onboard computation devices and an extensive corpus of perception data being generated from robots, also accelerates the adoption of learning methods for social robot navigation. 
\begin{figure}
    \centering
    \includegraphics[width=1\columnwidth]{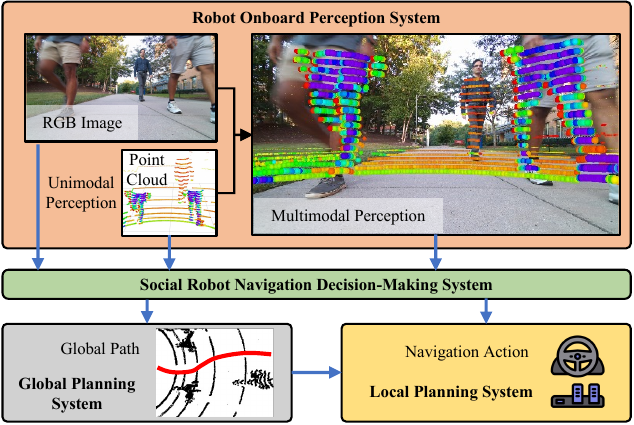}
    \caption{Social Robot Navigation Decision Making on the Global and Local Level with Multimodal and Unimodal (RGB Image and Point Cloud) Perception Input.}
    \label{fig::multimodal}
    \vspace{-20pt}
\end{figure}
Most current robots have multiple sensors onboard, with LiDARs and RGB cameras as the most common sensing modalities, and are therefore able to perceive complex social interactions from different sources (Fig.~\ref{fig::multimodal}). 
While LiDARs have been the main perception modality for mobile robots for decades, recent research has shifted towards visual navigation with RGB input alone, thanks to its cheap cost and wide availability. Intuitively speaking, LiDARs provide high-resolution and high-accuracy geometric information about the environments, while cameras stream in RGB images which contain rich semantics. Both geometric and semantic information play a role in the decision making process of social robot navigation: Geometric structures like obstacles and humans need to be avoided, while semantics including navigation terrain, human gaze~\cite{fiore2013toward, hart2020using}, gesture, clothing, and body language can shed light on the navigation contexts and other humans' intentions to inform robot navigation decisions. 

Considering the rich and potentially complementary information provided by multiple available sensor modalities onboard mobile robots and the efficiency of learning methods in enabling emergent social robot navigation behaviors, this paper presents a comprehensive study on using multimodal perception of LiDAR and RGB camera inputs, two most common perception modalities of autonomous mobile robots, to learn the robot decision making process during social robot navigation. The study is conducted on a large-scale real-world Socially Compliant Navigation Dataset (\textsc{scand})~\cite{karnan2022scand} collected in a variety of natural crowded public spaces on a university campus. From the social robot navigation perspective, we study the decision-making capability of multimodal and unimodal learning on both global and local planning in different social scenarios (e.g., Against Traffic, With Traffic, and Street Crossing); From the machine learning perspective, we study the training and generalizability performance of multimodal and unimodal learning in terms of training time, loss value, generalization accuracy, etc. We also conduct a human study and reveal how social compliance achieved by different sensor modalities can be perceived by humans interacting with the robot. The results show that multimodal learning 
is more reliable and robust 
than using
unimodal networks in both dataset and human studies.
\section{RELATED WORK}
\label{sec::related_work}
We review related work in social robot navigation, machine learning for navigation, and multimodal learning. 

\subsection{Social Robot Navigation}
While collision-free navigation has been investigated by the robotics community for decades~\cite{fox1997dynamic, quinlan1993elastic, perille2020benchmarking, nair2022dynabarn, xiao2022autonomous, xiao2023autonomous}, roboticists have also built mobile robots that navigate around humans since the early museum tour-guide robots RHINO~\cite{buhmann1995mobile} and MINERVA~\cite{thrun2000probabilistic}. Going beyond simply treating humans as dynamic, non-reactive obstacles~\cite{nair2022dynabarn}, researchers have also modeled the uncertainty of human movements~\cite{joseph2011bayesian, bennewitz2005learning, shiomi2014towards, unhelkar2015human, xu2022socialvae} or prescribed social norms for navigating agents~\cite{knepper2012pedestrian, sisbot2007human, luber2010people}, and then devised navigation planners that can take such uncertainty into account or abide such selected rules. These physics-based models~\cite{gupta2022intention, xu2021pfpn, xu2021human, truong2016towards} consider humans' behavior features, such as proxemics~\cite{hall1966hidden, kirby2009companion, takayama2011expressing, torta2013design}, intentions~\cite{dragan2013legibility, mavrogiannis2018social}, and social formations and spaces~\cite{vazquez2015social, vroon2015dynamics, fiore2013toward, shiomi2014towards, van2011reciprocal}. 
However, prescribing a simple model is usually not sufficient to capture complex human behaviors in the wild. For example, pedestrians move differently during rush hours or on weekends, within formal or informal contexts. 
Furthermore, such a plethora of factors to be considered during social robot navigation all have to be processed from raw perceptual data from onboard sensors, e.g., LiDARs and RGB cameras, and set challenges for onboard perception algorithms, e.g., human tracking, motion prediction, and intention detection. Along with the recent success in machine learning, both these challenges led to the recent adoption of data-driven approaches for social robot navigation~\cite{xiao2022motion}.

\subsection{Machine Learning for Navigation}
As a potential solution to the aforementioned challenges, machine learning approaches have been leveraged to implicitly encode the complexities and subtleties of human social behaviors in a data-driven manner~\cite{xiao2022motion} and also address other challenges in navigation, e.g., off-road navigation~\cite{xiao2021learning, karnan2022vi, atreya2022high, sikand2022visual, datar2023learning, datar2023toward}. These data-driven approaches include learning representations or costmaps~\cite{xiao2022learning, kim2016socially, kretzschmar2016socially, vasquez2014inverse, ziebart2009planning}, parameterizations of navigation planners~\cite{liang2021crowd, xiao2022appl, xiao2020appld, wang2021apple, wang2021appli, xu2021applr}, local planners~\cite{xiao2021toward, xiao2021agile, wang2021agile, francis2020long, xu2021machine}, or end-to-end navigation policies that map directly from raw or pre-processed perceptions of the humans in the scene to motor commands that drive the robot~\cite{chen2017socially, tai2018socially, pfeiffer2017perception}. From the perspective of machine learning methods, reinforcement learning~\cite{francis2020long, xu2023benchmarking, xu2021machine, xu2023benchmarking, xu2021applr} and imitation learning~\cite{nguyen2023toward, bojarski2016end, Nazeri2021, xiao2022learning, xiao2020appld, wang2021apple, wang2021appli} depend on training data from mostly simulated trial-and-error experiences and either human or artificial expert demonstrations respectively. Considering the difficulty in producing high-fidelity perceptual data and natural human-robot interactions in simulation, this study adopts an imitation learning setup, in particular, Behavior Cloning (BC)~\cite{bojarski2016end, Nazeri2021}, with a large-scale social robot navigation demonstration dataset. 

\subsection{Multimodal Learning}
Recent research has shown that combining data from different modalities in a multimodal learning framework can lead to promising results in solving downstream tasks~\cite{ramachandram2017deep}. 
For autonomous mobile robot navigation, researchers have tried sensor fusion by combining RGB cameras, LiDARs, and robot odometry with a multimodal graph neural network to navigate unstructured terrain including bushes, small trees, and grass regions of different heights and densities~\cite{weerakoon2023graspe}. Furthermore, they have demonstrated the robustness of the network towards partial occlusion and unreliable sensor information in challenging outdoor environments. Other researchers have also combined laser, RGB images, point cloud, and distance map to learn navigation in time-sensitive scenarios such as disaster response or search and rescue, which include constrained narrow passages, pathways with debris, and irregular navigation scenarios~\cite{nguyen2020autonomous}. Additionally, they have demonstrated that multimodal networks outperformed models that only utilized RGB images and distance maps. 
Multimodal perception has been shown to be valuable in addressing different challenges during real-world navigation tasks, but to the best of our knowledge, investigation into how multimodal perception can affect decision making during social robot navigation is still very limited, which is the focus of this study. Notice that we are interested in learning social robot navigation with multimodal \emph{perception} as input~\cite{ramachandram2017deep}, rather than learning models with multimodal \emph{distribution}, which has a relatively richer literature~\cite{srivastava2012multimodal, gupta2018social, li2020socially}. 


\subsection{Socially Compliant Robot Navigation Dataset (\textsc{scand})}
Our study is based on an open-source, large-scale, real-world social robot navigation dataset, \textsc{scand}~\cite{karnan2022scand}, of 8.7 hours, 138 trajectories, 40 kilometers of socially compliant, human teleoperated driving demonstrations that comprise multimodal data streams including 3D LiDAR, visual and inertial information, robot odometry, and joystick commands, collected on two morphologically different mobile robots---a Boston Dynamics Spot and a Clearpath Jackal---by four different human demonstrators in both indoor and outdoor environments. Due to its rich social interactions and multimodal perception-to-action navigation decisions, \textsc{scand} is suitable for studying social robot navigation learning with multimodal perception. Specifically, we study the effect of both point cloud data from a 3D LiDAR and RGB images from a camera, the most commonly available perception modalities onboard mobile robots, considering the geometric and semantic information provided by the point cloud data and RGB images can complement each other to assist decision making during social robot navigation in human-inhabited public spaces.

\section{MULTIMODAL LEARNING FOR SOCIAL ROBOT NAVIGATION}
\label{sec::approach}

\begin{figure*}
    \centering
    \includegraphics[width=2\columnwidth]{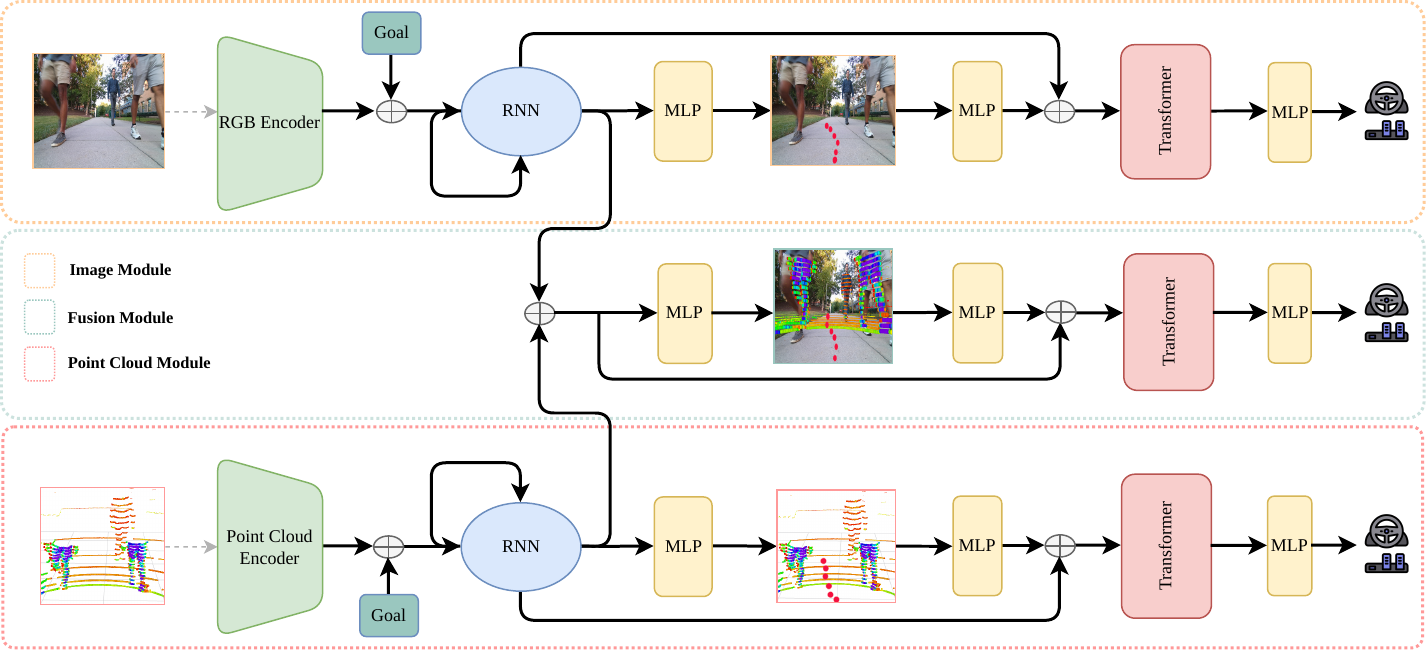}
    \caption{Image Module, Fusion (Multimodal) Module, and Point Cloud Module Architecture for Social Robot Navigation.}
    \label{fig::architecture}
    \vspace{-20pt}
\end{figure*}

We adopt an imitation learning approach, i.e., BC, to learn socially compliant navigation decisions using multimodal perception from \textsc{scand}. Similar to classical navigation systems with a global and a local planning system, we design our multimodal learning framework so that it will produce both global and local plans and study how multimodal and unimodal learning can imitate the navigation decisions made by the human demonstrator on both global and local levels. 

\subsection{Problem Formulation}
Specifically, at each time step $t$ of each trial in \textsc{scand}, the robot receives onboard perceptual input, including a sequence of 3D LiDAR point cloud data $L$ and RGB images $I$, and a goal $G$ it aims to reach, which is taken as a waypoint 2.5m away from the robot on the future robot odometry. We denote all these inputs necessary to inform the decision-making process during social robot navigation as a navigation input: $\mathcal{I}^{D}_t=\{L^{D}_k, I^{D}_k, G^{D}_t\}_{k=t-N+1}^t$, where $N$ denotes the history length included in the navigation input at $t$ and $D$ denotes that the data is from the \textsc{scand} demonstrations. 

Facing a social navigation input $\mathcal{I}^{D}_t$, the \textsc{scand} demonstrator shows the desired, socially compliant navigation decision $\mathcal{D}_t$ on both global and local levels: $P_t$ is the demonstrated global plan, recorded as the human-driven future robot odometry starting from time $t$, and takes the form of a sequence of 2D waypoints $P^{D}_t = \{(x^{D}_i, y^{D}_i)\}_{i=t}^{t+M-1}$; $A_t$ is the demonstrated local plan represented as a sequence of joystick action commands $A^{D}_t = \{(v^{D}_i, \omega^{D}_i)\}_{i=t}^{t+K-1}$, where $v$ and $\omega$ is the linear and angular velocity respectively. $M$ and $K$ denote the length of the navigation decision on the global and local plan level respectively. 
The demonstrated navigation decision is therefore defined as $\mathcal{D}^{D}_t = \{P_t^D, A_t^D\}$. 

Producing the navigation decision  $\mathcal{D}^{D}_t$ based on $\mathcal{I}_t^D$ as input, a navigation system is defined as a combination of two functions, $\mathcal{F}^g(\cdot)$ and $\mathcal{F}^l(\cdot)$, responsible of generating the global plan $P_t^D$ and local plan (action) $A_t^D$: \
\begin{equation}
\begin{split}
    P_t^D &=  \mathcal{F}^g(\mathcal{I}^{D}_t),\\
    A_t^D &= \mathcal{F}^l(\mathcal{I}^{D}_t, P_t^D). 
    \nonumber
\end{split}
\end{equation}
In a data-driven manner, we instantiate both global and local planners by learning $\mathcal{F}^g_\theta(\cdot)$ and $\mathcal{F}^l_\phi(\cdot)$ as deep neural networks with learnable parameters $\theta$ and $\phi$ respectively. In particular, we aim to learn the parameters to minimize a BC loss: 
\begin{equation}
\begin{gathered}
    \theta^*, \phi^* = \argmin_{\theta, \phi} \sum_{P_t^D, A_t^D, \mathcal{I}_t^D \in \textsc{scand}}\\
    \left[ ||P_t^D - \mathcal{F}^g_\theta(\mathcal{I}^{D}_t)|| + \lambda ||A_t^D - \mathcal{F}^l_\phi(\mathcal{I}^{D}_t, \mathcal{F}^g_\theta(\mathcal{I}^{D}_t))|| \right],
\end{gathered}
\label{eqn::loss}
\end{equation}
where the first term is the difference between demonstrated and learned global plan, while the second term is for the local plan, with $\lambda$ as a weight between them. 

In this study, we are interested in studying the effect of including different perception modalities in $\mathcal{I}_t$ on making socially compliant navigation decisions $P_t$ and $A_t$. We study three scenarios, i.e., multimodal perception $\mathcal{I}_t^\textrm{MM} = \{L_k, I_k, G_t\}_{k=t-N+1}^t$, unimodal LiDAR (point cloud) perception $\mathcal{I}_t^\textrm{LiDAR} = \{L_k, G_t\}_{k=t-N+1}^t$, and unimodal vision (RGB image) perception $\mathcal{I}_t^\textrm{Vision} = \{I_k, G_t\}_{k=t-N+1}^t$. For simplicity and consistency, we keep $N=1$ for all three cases in this study and leave an investigation into different history lengths as future work. 

\subsection{Unimodal Perception}
\subsubsection{Point Cloud Modality}
We take points that are within the range of 8 meters in front, 3 meters on either side and within 2.5 meters of height from the robot as perceived by the 3D LiDAR. All points are placed into their respective voxel inside a 3D voxel grid with $5\times5\times5$cm voxels, resulting in a $160\times120\times50$ voxel representation for $L_k$. 
We use a 3D Convolution Neural Network (CNN)~\cite{maturana2015voxnet} to process the voxel representation to extract meaningful information for our downstream social robot navigation task. The point cloud encoder is shown as the green trapezoid in  the red box at the bottom of Fig.~\ref{fig::architecture}. 

\subsubsection{RGB Modality}
For RGB images, we take a $224\times224\times3$ image from the camera as input. 
We use ResNet-18~\cite{he2016deep} to extract features for our social robot navigation task. The image encoder is shown as the green trapezoid in the yellow box at the top of Fig.~\ref{fig::architecture}. 

Both RGB and point cloud inputs have their own unimodal decision making modules, shown in the upper yellow and lower red box in Fig.~\ref{fig::architecture} respectively. For a fair comparison, we enforce the same architecture, the only difference is the different input modalities. To be specific, we concatenate the embeddings from the corresponding input encoders with the local goal (2.5m away), and feed them into a Recurrent Neural Network (RNN) to capture history information (blue ellipsoids in Fig.~\ref{fig::architecture}). Then we use a Multi-Layer Perceptron (MLP) (yellow boxes in Fig.~\ref{fig::architecture}) to produce global plan in the form of a sequence of 2D waypoints (red dots in Fig.~\ref{fig::architecture}), which are further fed into another MLP. Concatenating the MLP output with the RNN output, a transformer, and another MLP at the end produces local plan, i.e., actions of linear and angular velocities. 

\subsection{Multimodal Fusion}
For multimodal fusion, the outputs of the RNNs from the point cloud and image modules are concatenated and passed through the fusion process, shown in Fig.~\ref{fig::architecture} middle. Similar to the unimodal modules, our feature fusion also happens at two different places in our multimodal network. Each fusion caters to different downstream tasks, i.e., producing both global and local plans.  

\subsection{Navigation Decisions and Loss Functions}
The global navigation decisions are instantiated as a sequence of five future waypoints ahead of the robot, i.e., $P^{D}_t = \{(x^{D}_i, y^{D}_i)\}_{i=t}^{t+4}$ ($M=5$), each of which is 0.5m apart taken from the future robot odometry. The local navigation decisions take the form of the current linear and angular velocity commands, i.e., $A^{D}_t = \{(v^{D}_t, \omega^{D}_t)\}$  ($K=1$). 

For the first and second loss terms in Eqn.~\ref{eqn::loss}, we use $L2$-norm of the five future waypoints and $L1$-norm of the current angular and linear velocity. We set $\lambda=1$. 

\subsection{Design Choices}
Notice that all aforementioned design choices with respect to neural network hyper-parameters and architecture are made after extensive trial-and-error and careful fine-tuning to ensure the different modalities can achieve the best learning performance for a fair comparison. All detailed hyper-parameters and design choices can be found in our open-source implementation for the future use of the community. 

We have experimented with PointNet~\cite{qi2017pointnet} and PointNet++~\cite{qi2017pointnet++} for the point cloud encoder, which does not perform well on \textsc{scand} social navigation scenarios: PointNet encodes individual point and relies on the global pooling layers to extract effective features. However, encoding points for highly diverse indoor and outdoor \textsc{scand} scenarios is not effective. Unlike closed, and small-scale indoor objects, point clouds collected during real-world robot navigation contain significantly more variation in terms of the number and distribution of points.
Our further investigation into the point cloud encoder reveals that converting them to a voxelized grid and then processing them through a 3D CNN network results in a significant performance gain. 

We also try to learn local planner using simple MLP, but it fails to capture the variations in \textsc{scand}. For instance, for the same global path, there can be different velocities: If humans are nearby the linear velocity will be slower, in contrast to a scenario where they are far apart. 
Transformer can achieve significant performance gain because of the attention modules which can decide which features it should attend to in order to  capture these variations.


\section{SCAND STUDY RESULTS}
\label{sec::results}

\begin{figure*}
    \centering
    \includegraphics[width=2\columnwidth]{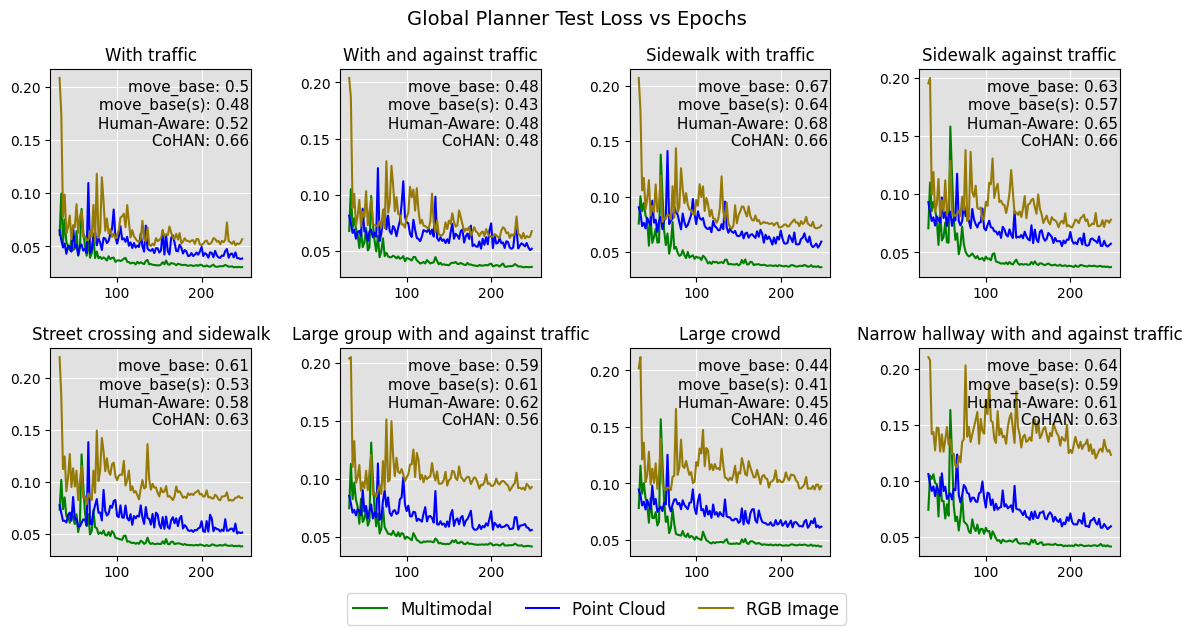}
    \caption{Test Loss on Eight \textsc{scand} \textsc{rosbags} with Multimodal, Point Cloud, and RGB Image Input (Averaged Over Three Training Runs with Negligible Variances Invisible in the Figures). }
    \label{fig::results}
    \vspace{-20pt}
\end{figure*}

We first present our study results on all the social scenarios in \textsc{scand} before presenting our human study results. We divide the \textsc{scand} trials into 18 for training and 8 for testing. We analyze the learning results on the test data from both the machine learning and social robot navigation perspectives. 
The training loss curves for the global planner in terms of L1 loss on the eight \textsc{scand} \textsc{rosbags} are shown in Fig.~\ref{fig::results}, while the local planner loss in Fig.~\ref{fig::local_planner_results}. We also plot the performance of a variety of classical social robot navigation planners using the same loss function between their output and the \textsc{scand} demonstration to compare against end-to-end learned policies.

\subsection{Multimodal Learning Performance}
The results of the eight test \textsc{scand} \textsc{rosbags} are ordered roughly according to increasing performance discrepancy among different modalities in Fig.~\ref{fig::results}, which can also be treated as an approximate representation of the ``difficulty'' level in social robot navigation decision making. For example, the loss values of most modalities converge faster and to a lower point in the earlier ``easy'' trials (upper left), compared to the later ``difficult'' ones (lower right). 

It is clear that in terms of test loss for global planning, learning with multimodal perception significantly outperforms both unimodal perception modalities. The multimodal test loss shown by the green curves drops faster, converges at a smaller epoch number, and reduces to a lower value compared to both the yellow and blue curves for RGB image and point cloud respectively. It is also worth noticing that the green multimodal learning curves are similar and consistent across all eight  test \textsc{scand} \textsc{rosbags} with different social interactions in different social scenarios, showing the advantage of multimodal learning from both point cloud and RGB image. 

Another very clear trend is that for the two unimodal perception types, point cloud perception consistently outperforms RGB image in all test trials, despite underperforming multimodal learning. In the earlier ``easier'' trials, point cloud performs slightly better than RGB image and has a relatively larger discrepancy compared to multimodal learning. For the later ``difficult'' trials, such trend is reversed, with the point cloud blue curves come closer to the multimodal green curves, compared to the RGB yellow curves. 

\begin{figure}
    \centering
    \includegraphics[width=0.8\columnwidth]{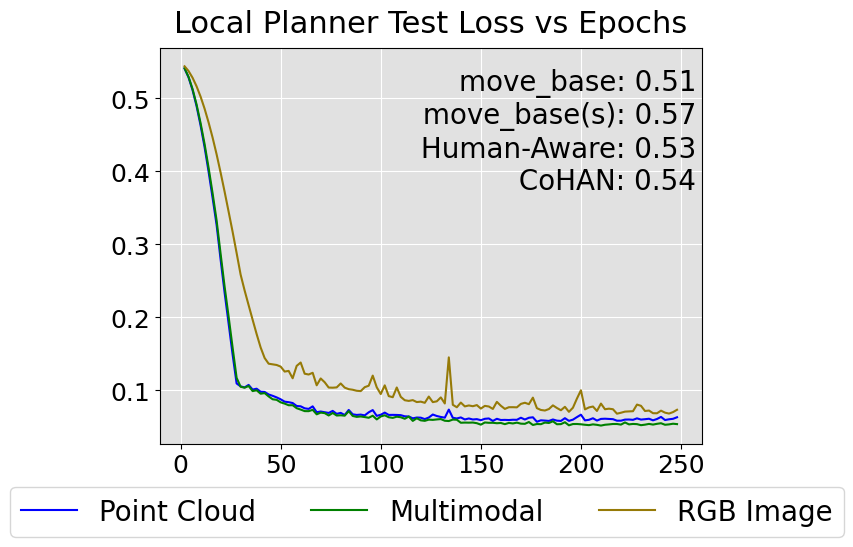}
    \caption{Average Test Loss on All \textsc{scand} \textsc{rosbags} with Multimodal, Point Cloud, and RGB Image Input (Three Training Runs).}
    \label{fig::local_planner_results}
\end{figure}

Considering that there is no significant difference on the local planning loss curves across the eight test \textsc{scand} \textsc{rosbags}, for the sake of space, we combine all eight curves into one for each modality and show them in Fig.~\ref{fig::local_planner_results}. We observe a similar trend in learning local planning from all three perception modalities: Multimodal learning can achieve 
slightly better performance at imitating the \textsc{scand} demonstrations than learning with the point cloud, which further outperforms learning with RGB image.

\subsection{Multimodal Social Compliance}
In addition to the pure machine learning statistics, we also discuss how each perception modality performs with different social interactions in different social scenarios. As discussed above, the learning performance of RGB image decreases from the first to the last test trial and results in a more-than-doubled loss value in Fig.~\ref{fig::results}, while multimodal learning and point cloud learning consistently maintain similar performance. We also list the majority of the social scenarios presented in each test \textsc{scand} \textsc{rosbag} at the top of each subfigure in Fig.~\ref{fig::results}. We observe that the increasing ``difficulty'' level (mostly for RGB images) directly corresponds to increased human density caused by more confined social spaces and larger number of humans in the crowd. While learning with RGB image produces performance only slightly worse than point cloud and multimodal learning in the first ``with traffic'' scenario, which is a relatively simple scenario on a wide open walkway on the UT Austin campus, including ``against traffic'' human crowds and constraining navigation on a sidewalk instead of an open walkway deteriorates the performance of learning with RGB image only (first rown in Fig.~\ref{fig::results}). When the ``difficulty'' level keeps increasing by adding more complex social scenarios such as ``street crossing'', ``large group/crowd'', and ``narrow hallway'', RGB image's performance keeps degrading. We posit that such performance degradation is caused by the increased complexity and variance in the RGB input, which prevent learning with RGB image only from generalizing to unseen data in challenging social scenarios. Furthermore, considering the lack of direct and explicit geometric information from RGB images, operating mobile robots in confined social spaces with large human crowds is also less safe compared to point cloud, whose geometric information can be utilized to assure safety, i.e., asserting a safe stopping behavior when the distance between the robot and the humans in the scene is too close. Such a lack of safety by relying only on RGB images is also apparent in our human study (see details in Sec.~\ref{sec::human}). 

The obvious gap between multimodal and point cloud learning is also of interest. While both of them are able to perform similarly across all eight test \textsc{scand} \textsc{rosbags}, multimodal learning maintains a very consistent advantage over point cloud alone in terms of a lower converged loss value and fewer epochs until convergence. We posit that the additional semantic information provided by the RGB image in addition to the pure geometric data from point cloud can provide extra relevant social cues to inform social navigation decision making. Such an empirical gap reveals the necessity of including semantic information in the social robot navigation decision making process, compared to traditional autonomous mobile robot navigation, for which avoiding obstacles is the only concern.

\section{HUMAN STUDY RESULTS}
\label{sec::human}
We conduct a human study to test whether the findings from our \textsc{scand} study can translate to real-world social robot navigation. We use a Clearpath Jackal robot with a Velodyne VLP-16 LiDAR and a ZED2 RGB-D camera for the point cloud and RGB image input respectively. We recruit eight human subjects for our human study. 

\begin{figure}
    \centering    \includegraphics[width=1\columnwidth]{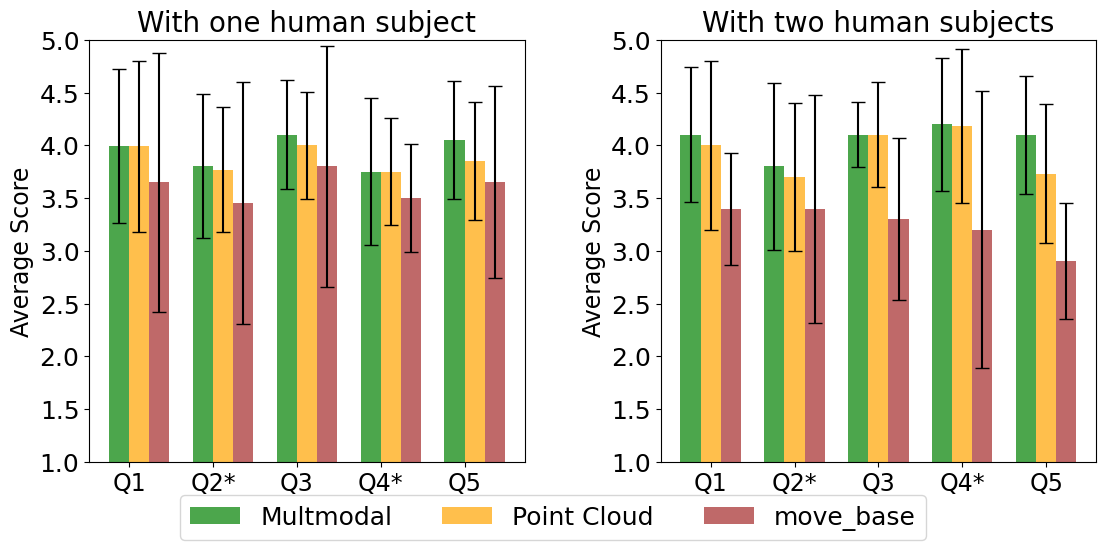}
    \caption{Human Study Results.}
    \label{fig::human}
    \vspace{-20pt}
\end{figure}

\begin{figure}
    \centering
    \includegraphics[width=1\columnwidth]{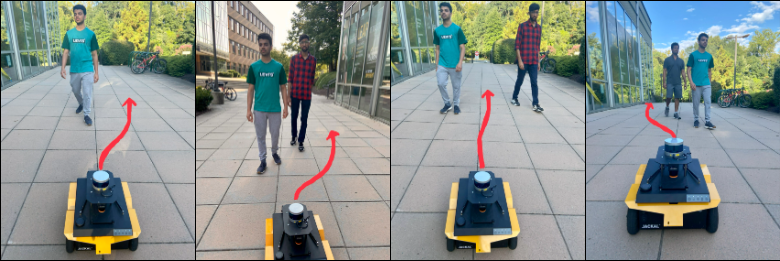}
    \caption{Human Study with Different Social Scenarios.}
    \label{fig::human_study_photos}
    \vspace{-20pt}
\end{figure}

Two sets of experiments are designed according to a previous protocol to evaluate social robot navigation~\cite{pirk2022protocol}: frontal approach of the robot with one and two human participants in a public outdoor space (Fig.~\ref{fig::human_study_photos}). In the one-human study, participants are instructed to take a natural path towards the robot; Participants in the two-human study are instructed to take three different approaches to initiate social interactions: move directly towards the robot, move forward then diverge, and move towards one side of the robot. After deploying the RGB module, we found that the robot may move dangerously close to the human subjects. Therefore, we exclude the RGB module in the human study. 

After each human-robot interaction, we ask the participant to fill in a standard questionnaire~\cite{pirk2022protocol} with five questions\footnote{$^*$ denotes negatively formulated questions, for which we reverse-code the ratings to make them comparable to the positively formulated ones.}: \emph{1. The robot moved to avoid me}, \emph{2. The robot obstructed my path$^*$}, \emph{3. The robot maintained a safe and comfortable distance at all times}, \emph{4. The robot nearly collided with me$^*$}, and \emph{5. It was clear what the robot wanted to do}.


The per-question average along with error bars are plotted in Fig.~\ref{fig::human} for both the one-person (left) and two-person scenarios (right). For all five questions, the multimodal learning approach is able to consistently achieve higher social compliance scores with smaller variance, compared to \texttt{move\_base}, the best classical planner according to the loss values in the \textsc{scand} study. Compare the left and right figures, the difference between multimodal learning and \texttt{move\_base} increases with more humans, showing multimodal learning's potential to enable socially compliant navigation with higher human density in public spaces, which is consistent with the results we observe in terms of test loss values in the \textsc{scand} study (Fig.~\ref{fig::results}). For our curated human study, we do not observe a significant advantage of multimodal learning in comparison to point cloud only. We posit that it is because our curated social scenarios do not contain sufficiently rich semantic social cues to showcase the necessity of using RGB images.

\section{CONCLUSIONS}
\label{sec::conclusions}

We present a study on learning social robot navigation with multimodal (and unimodal) perception conducted on both a large-scale real-world social robot navigation dataset and in a human study with a physical robot, in comparison to a set of classical approaches. Our study results indicate that multimodal learning has clear advantage over either unimodal counterpart by a large margin in both the dataset and human studies, especially in difficult situations with increasing human density. In terms of unimodal learning, point cloud input is superior compared to RGB input, but it can be improved by utilizing the extra semantic information provided by the camera. Despite the found superiority of multimodal learning, the current study only remains in pre-recorded dataset and curated social scenarios. How multimodal learning will perform in real-world, large-scale, long-term social robot navigation tasks remains unclear and may require extra research and engineering effort.

\bibliographystyle{IEEEtran}
\bibliography{IEEEabrv,references}
\end{document}